\documentclass{article}

\usepackage{arydshln}
\usepackage{microtype}
\usepackage{graphicx}
\usepackage{subfigure}
\usepackage{float}
\usepackage{booktabs} 

\usepackage[most]{tcolorbox}
\newtcolorbox{myquote}[1][]{%
    colback=black!3,
    colframe=black!3,
    notitle,
    sharp corners,
    borderline west={2pt}{0pt}{blue!80!black},
    enhanced,
    breakable,
}

\usepackage{hyperref}

\newcommand{\etc}{\emph{etc.}}

\newcommand{\tablestyle}[2]{\setlength{\tabcolsep}{#1}\renewcommand{\arraystretch}{#2}\centering\footnotesize}
\newlength\savewidth\newcommand\shline{\noalign{\global\savewidth\arrayrulewidth
		\global\arrayrulewidth .8pt}\hline\noalign{\global\arrayrulewidth\savewidth}}		
\newif\ifshowcomment
\showcommentfalse

\newcommand{\nllm}{14 }

\usepackage[accepted]{icml2023}
\usepackage{footmisc}
\usepackage{amsmath}
\usepackage{amssymb}
\usepackage{mathtools}
\usepackage{amsthm}
\usepackage{enumitem}
\usepackage[ruled,linesnumbered]{algorithm2e}
\usepackage[capitalize,noabbrev]{cleveref}

\theoremstyle{plain}

\theoremstyle{definition}

\theoremstyle{remark}

\usepackage[textsize=tiny]{todonotes}

\icmltitlerunning{}

\begin{document}

\twocolumn[
\icmltitle{BotChat: Evaluating LLMs' Capabilities of Having Multi-Turn Dialogues}



\icmlsetsymbol{equal}{*}
\icmlsetsymbol{lead}{+}

\begin{icmlauthorlist}
\icmlauthor{Haodong Duan}{equal,lead,sai}
\icmlauthor{Jueqi Wei}{equal,sai,fdu}
\icmlauthor{Chonghua Wang}{sai,sjtu}
\icmlauthor{Hongwei Liu}{sai,fdu} \\
\icmlauthor{Yixiao Fang}{sai}
\icmlauthor{Songyang Zhang}{sai}
\icmlauthor{Dahua Lin}{sai,cuhk}
\icmlauthor{Kai Chen}{sai}
\end{icmlauthorlist}

\icmlaffiliation{sai}{Shanghai AI Laboratory}
\icmlaffiliation{fdu}{Fudan University}
\icmlaffiliation{sjtu}{Shanghai Jiaotong University}
\icmlaffiliation{cuhk}{The Chinese University of Hong Kong}

\icmlcorrespondingauthor{Kai Chen}{chenkai@pjlab.org.cn}


\vskip 0.3in
]



\printAffiliationsAndNotice{\icmlEqualContribution} 

\begin{abstract}

Interacting with human via high-quality multi-turn dialogues is a key feature of large language models (LLMs). 
However, human-based evaluation of such capability involves intensive manual labor. 
This report provides a preliminary evaluation of existing large language models 
for human-style multi-turn chatting, through an LLM-based approach.
We start from real-world human dialogues and keep the very first utterances as the ChatSEED.
Then we prompt LLMs to generate a full multi-turn dialogue (tens of utterances) based on the ChatSEED, utterance by utterance. 
Finally, we adopt state-of-the-art LLMs (GPT-4, \etc) as the judge to evaluate the generated dialogues.
With different evaluation protocols, we come to substantially identical conclusions.
We find that GPT-4 can generate human-style multi-turn dialogues with impressive quality, significantly outperforms its counterparts.
It's difficult for a discriminator to distinguish between GPT-4 generated dialogues and human dialogues. 
In contrast, other LLMs struggle to generate multi-turn dialogues of satisfactory quality due to
poor instruction-following capability, tendency to generate lengthy utterances, or limited general capability.
All data and codes will be provided in \url{https://github.com/open-compass/BotChat/} and we hope they can serve as a valuable resource for evaluating multi-turn chatting capabilities of LLMs.

\end{abstract}
\section{Introduction}

\begin{figure}[t]
\centering
\includegraphics[width=\linewidth]{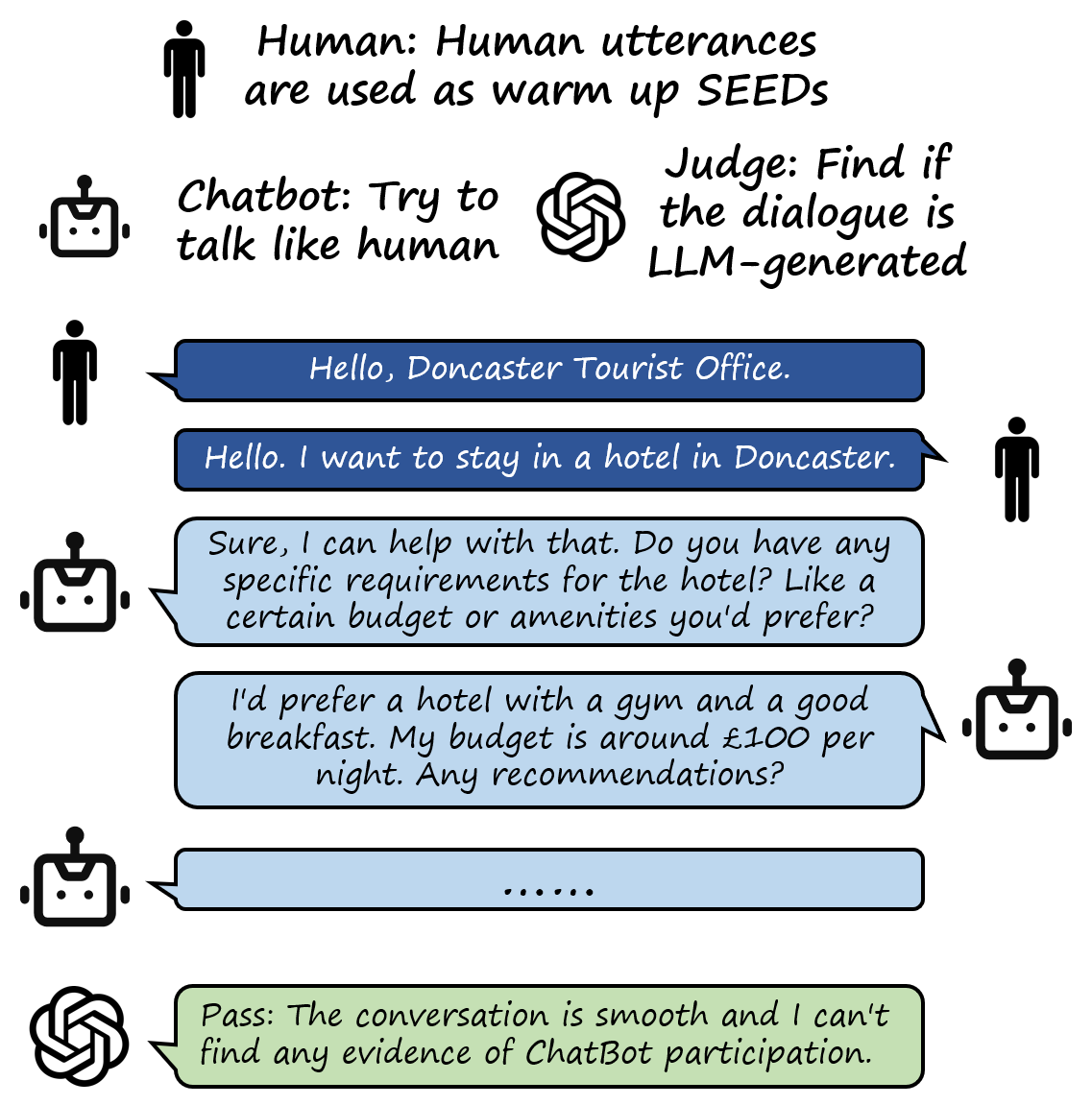}
\vspace{-7mm}
\caption{\textbf{BotChat} evaluates the multi-turn dialogue performance by prompting ChatBots to generate multi-turn dialogues based on initial human utterances and evaluate them with a judge LLM. }
\label{fig:teaser}
\vspace{-6mm}
\end{figure}

The recent progress of Large Language Models (LLMs) \citep{openai2023gpt4,touvron2023llama,chiang2023vicuna} represents a significant advancement in artificial intelligence, and has a profound impact.
Compared to traditional language models~\citep{devlin2018bert,vaswani2017attention,liu2019roberta}, 
modern LLMs can chat with human much better. 
Specifically, LLMs can interact with people with smooth multi-turn conversations in the human style, 
learn the instruction, intention, and context from human prompts to provide helpful feedback. 
Such advantage enables all human kinds to directly access the strong capability of LLMs for various applications, either general~\citep{jiao2023chatgpt,shen2023hugginggpt} or within specific domains~\citep{bran2023chemcrow,boiko2023emergent}. 

Fluently engaging in multi-turn conversations with humans is a fundamental capability of modern LLMs. 
However, not all models excel in this domain. 
In practical applications, it has been observed that dialogues generated by certain LLMs frequently fail to meet user satisfaction criteria. 
The issues manifest in multiple aspects, including poor adherence to user instructions, undesirable tone or lengthy utterances, and the generation of repetitive content.
The assessment of this multi-turn conversational ability remains an open and challenging problem.
The most commonly adopted approach is based on human~\citep{zheng2023judging},
which intensively involves manual labor for human-bot conversation generation and conversation quality assessment.
In this project, we propose a more efficient paradigm, named \textbf{BotChat}, to evaluate the multi-turn chatting capability. 

BotChat presents a pure LLM-based solution for multi-turn conversational capability evaluation, 
with no manual labor required. 
The process includes two stages: dialogue generation and quality assessment. 
In stage 1, we start from the very first utterances (ChatSEED) from real-world conversations~\citep{cui2020mutual} and conduct \textbf{utterance-by-utterance} generation.
In each step, a ChatBot generates one utterance based on all history utterances.
With this strategy, we automatically generates multi-turn dialogues at arbitrary lengths.

In stage 2, we evaluate the quality of generated conversations with multiple LLM-based evaluation protocols, 
for cross validation. 
We propose three evaluation protocols:
1. unitary evaluation (\textbf{UniEval}): evaluating each generated dialogue separately; 
2. pairwise evaluation (\textbf{BotChat Arena}): comparing two generated dialogues and determine the one with better quality; 
3. ground-truth evaluation (\textbf{GTEval}): comparing the generate dialogue with the corresponding human dialogue, and determine which one is real-human conversation. 
UniEval and BotChat Arena apply to conversations with arbitrary turns, while GTEval is confined by the turns of the number of turns of the ground-turth dialogue.

With the evaluation protocols, 
we compare \nllm representative LLMs, 
ranging from the state-of-the-art closed-source GPT-4~\citep{openai2023gpt4} to small-scale open-source LLMs~\citep{touvron2023llama2,bai2023qwen}.
During evaluation, the three evaluation protocols draw substantially identical conclusions.
GPT-4 generates human-style multi-turn conversations with impressive quality, outperforming all other LLMs.
For all LLMs, the quality of generated dialogues declined quickly as long as the dialogue turns increase. 
Such degradation is particularly evident for open-source LLMs at small scale, compared to the top-tier LLM GPT-4.
With qualitative assessment, 
we find that LLMs fail to generate multi-turn conversations with desirable quality primarily due to: 
poor instruction-following capability, tendency to generate lengthy utterances, and limited general capability.
\section{Related Works}

\subsection{Objective Assessment of LLMs}

Objective assessment of LLMs aims to evaluate the capabilities of LLMs in an objective and quantitative manner, usually achieved by comparing LLMs' outputs with  the corresponding reference or ground truth. 
For close-ended tasks~\citep{huang2023ceval,hendrycks2020measuring}, the outputs of LLMs are expected to be exactly the same with ground truths. 
For open-ended tasks~\citep{huang2021efficient,fabbri2019multi}, similarity metrics are calculated based on the outputs and references.
Higher similarity indicates a better performance on the corresponding task.
Several metrics including F1-score, BLEU~\citep{papineni2002bleu}, and ROUGE~\citep{lin2004rouge} are widely used in the measurement of their performance.

In BotChat, the conversation generation stage can be viewed as an open-ended natural language generation (NLG) task.
Although there exists a reference conversation for each generated conversation, 
the similarity-based objective assessment makes no sense~\citep{liu2016not}.
The reason is that even given identical ChatSEEDs, the following conversation may take various directions. 
Therefore, even though a generated conversation may be entirely different from a reference conversation, 
it does not necessarily indicate poor quality of the generated dialogue.


\subsection{Subjective Assessment of LLMs}
In order to better evaluate LLMs in more complicated scenarios, 
subjective assessment has been applied in many works~\citep{xu2023wizardlm,chiang2023vicuna} to compare the performance of different LLMs by humans or other powerful LLM judges. 

With the development of LLMs, some of powerful LLMs are employed as a surrogate to humans in measurement. 
GPTScore~\citep{fu2023gptscore} excavates the ability of GPT3 in achieving customized, multi-aspect and training-free evaluation. 
\cite{wang2023chatgpt} introduces a preliminary meta-evaluation on the reliability of ChatGPT as a judge. 
Chatbot Arena~\citep{zheng2023judging} is constructed for battles between chatbots based on both LLMs and humans. It further showcases high agreement between GPT-4 and humans on making judgements. 

When it comes to generated conversations, human-based evaluation leverages human preference to evaluate chitchat dialogue~\citep{dinan2018wizard,zhang2018personalizing,adiwardana2020towards,dougruoz2022open}. 
\cite{li2019acute} proposes an evaluation method, which asks human judges to make a binary choice between two models. 
However, manual evaluation suffers from high expense and low efficiency in measuring a large amount of cases.
To our best knowledge, we first adopt the powerful LLM as the judge to assess the quality of generated conversations. 
The LLM-based evaluation is more efficient compared to the human-based one, and can also yield robust yet reliable evaluation results. 

\subsection{Human Conversation Datasets}
\cite{serban2015survey} provides a list of conversation datasets for building end-to-end chatbots. 
PERSONA-CHAT~\citep{zhang2018personalizing} is an engaging dialogue dataset which displays consistent personality. 
\cite{zhou2018dataset} introduces specified document into multi-turn conversations. 
CoQA~\citep{reddy2019coqa} is a conversational question answering dataset collected by two annotators chatting about a passage. 
MuTual~\citep{cui2020mutual} consists of dialogues based on Chinese student English listening comprehension exams, aiming at facilitating conversation model reasoning capabilities.
In BotChat, we adopt MuTual as our major data source of human dialogues.
\section{BotChat}
In this section, we introduce the evaluation paradigms adopted in BotChat, which is built upon generated multi-turn dialogues. 
The framework purposes to assess the conversational performance of LLM concerning human subjective preferences. 
We first provide an overview of the workflow for generating multi-turn dialogues.
Subsequently, we introduce the three evaluation strategies proposed: unitary evaluation (\textbf{UniEval}), pairwise evaluation (\textbf{BotChat Arena}), and ground-truth evaluation (\textbf{GTEval}).

\subsection{Dialogue Generation}
In BotChat, we generate multi-turn dialogues solely based on ChatBots. 
To achieve this, we start with real world dialogues, and extract ChatSEEDs (the first several utterances) from them to serve as the generation condition.  
We resort to the existing dataset MuTual~\citep{cui2020mutual} for real world human conversations. 
We employ the first two utterances of each dialogue in its test split as the ChatSEED for dialogue generation. 
When generating the subsequent utterance within a dialogue, 
we first provide a system prompt\footnote{The prompt is provided in \Cref{prompt:dlg_gen}. \label{footnote:prompt}} to the LLM to encourage it to generate human-style utterances with relatively short lengths.  
The generating process follows an \textbf{utterance-by-utterance} manner. 
In each turn, we provide the system prompt as well as all previous utterances in the dialogue to the ChatBot, 
and the Chatbot generates the next utterance.
The process is repeated until we reach the target turns. 
We provide the PseudoCode for our generation paradigm in \Cref{alg:dlg_gen}.

\vspace{-3mm}
\begin{algorithm}
\SetAlgoLined
\KwData{ChatSEED $s$ (a list of two utterances); target number of rounds $N$; system prompt $SYS$; ChatBot $M$}
\KwResult{Generated Dialogue $D$ (a list of utterances)}
$D \leftarrow s$\;

$T \leftarrow len(D)$\;

\While{$T < N$}{
    $History \leftarrow$ build\_history($SYS$, $D$[:-1])\;
    
    $Utterance \leftarrow$ $M$.chat($D$[-1], $History$)\;
    
    $D$.append($Utterance$)\;

    $T \leftarrow len(D)$\;
}
    \caption{LLM-based Dialogue Generation. }
    \label{alg:dlg_gen}
\end{algorithm}

\subsection{UniEval}

One of the most straightforward evaluation strategies is independently evaluating each generated dialogue to determine its quality and similarity to human dialogues.  
We name this approach \textbf{UniEval}. 
Our structured evaluation encompasses the following steps:

\begin{enumerate}[leftmargin=*, itemsep=-.5ex, topsep=-1ex]
\item We first ask the judge LLM whether it perceives the provided dialogue as a ChatBot participated one (Y/N).
\item Depending on the response of the judge LLM:
\begin{enumerate}[leftmargin=*, itemsep=-.5ex, topsep=-1ex]
    \item If the judge LLM responds ``Yes", we request it to identify the index of the first utterance it deems to be generated by a ChatBot.
    \item If the judge LLM response ``No", we move forward without further inquiry.
\end{enumerate}
\item In conclusion, the LLM judge needs to provide an explanation for its judgment, offering valuable insights into the model's decision-making process.
\end{enumerate}

We also prepared several in-context examples that will be appended to the evaluation prompt\footref{footnote:prompt}, to strengthen the instruction following the capability of GPT-4.

\subsection{BotChat Arena}
UniEval has provided us with some initial insights. 
Nevertheless, it's important to recognize its inherent limitations. 
Despite our best efforts in providing evaluation guidance and contextual examples for GPT-4 evaluators, 
defining an explicit decision boundary to distinguish human conversations from LLM-generated ones remains a difficult challenge.

An alternative and widely adopted approach for benchmarking LLMs involves a comparative evaluation of responses from two distinct models when presented with the same question / message. 
This approach employs either humans or GPT-4 as evaluators. 
A notable benchmark following this paradigm is Chatbot Arena~\citep{zheng2023judging}, 
in which users interact with two separate LLM instances and prompt them with the same message or question. 
After the two responses are generated, the user further perform the assessment and select the preferable one.
The benchmark host gathers the feedback from a diverse group of users, and calculate an overall ELO rating~\citep{elo1967proposed} for each LLM.

Drawing inspiration from this paradigm, we introduce another evaluation strategy within the scope of our methodology, named ``\textbf{BotChat Arena}". 
In BotChat Arena, we employ a judge LLM to compare two conversations and determine whether the presented dialogues are ChatBot generated.
All the dialogues are pre-generated, without a chronological order.
To ensure a robust evaluation result and avoid the negative impact of randomness, 
we calculate the Bootstrap ELO, instead of the vanilla ELO rating.
Specifically, we shuffle the comparisons and calculate the vanilla ELO $N=1000$ times.
and report the median ELO rating for each LLM.
Experiment results show that the Bootstrap ELO is a stable metric across different random seeds.

\begin{figure}[t]
    \centering
    \includegraphics[width=.7\linewidth]{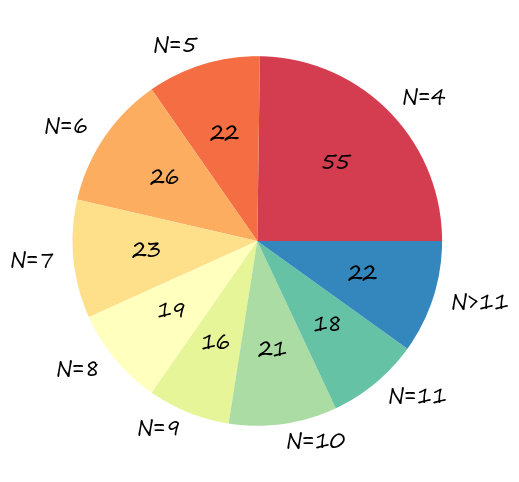}
    \vspace{-4mm}
    \caption{\textbf{Distribution of dialogue rounds in MuTual test.}}
    \label{fig:round_distribution}
    \vspace{-4mm}
\end{figure}

\subsection{GTEval}
``GTEval" involves a comprehensive comparison of the generated conversations with the ``Ground Truth" conversations in the test split of MuTual~\citep{cui2020mutual}. 
We adopt a similar protocol as employed in BotChat Arena for this evaluation. 
GTEval allows us to rigorously assess how well language models align with real human interactions, leveraging the rich resources of human dialogues.

To facilitate this comparison, we select a subset of 222 conversations from MuTual-Test, with each conversation comprises at least 4 utterances. 
The distribution of the conversation turns is listed in \Cref{fig:round_distribution}. 
Given that Ground Truth conversations may exhibit varying lengths, ranging from 4 to 15 chats, to ensure a fair and accurate comparison, we truncate all generated conversations to match the length of the reference Ground Truth conversation.
The meta prompt used in GTEval closely aligns with the one utilized in BotChat Arena, with a minor distinction. 
In this case, we explicitly specify that only one of the two dialogues contains LLM-generated utterances.
\begin{figure*}[ht]
\centering
\includegraphics[width=\linewidth]{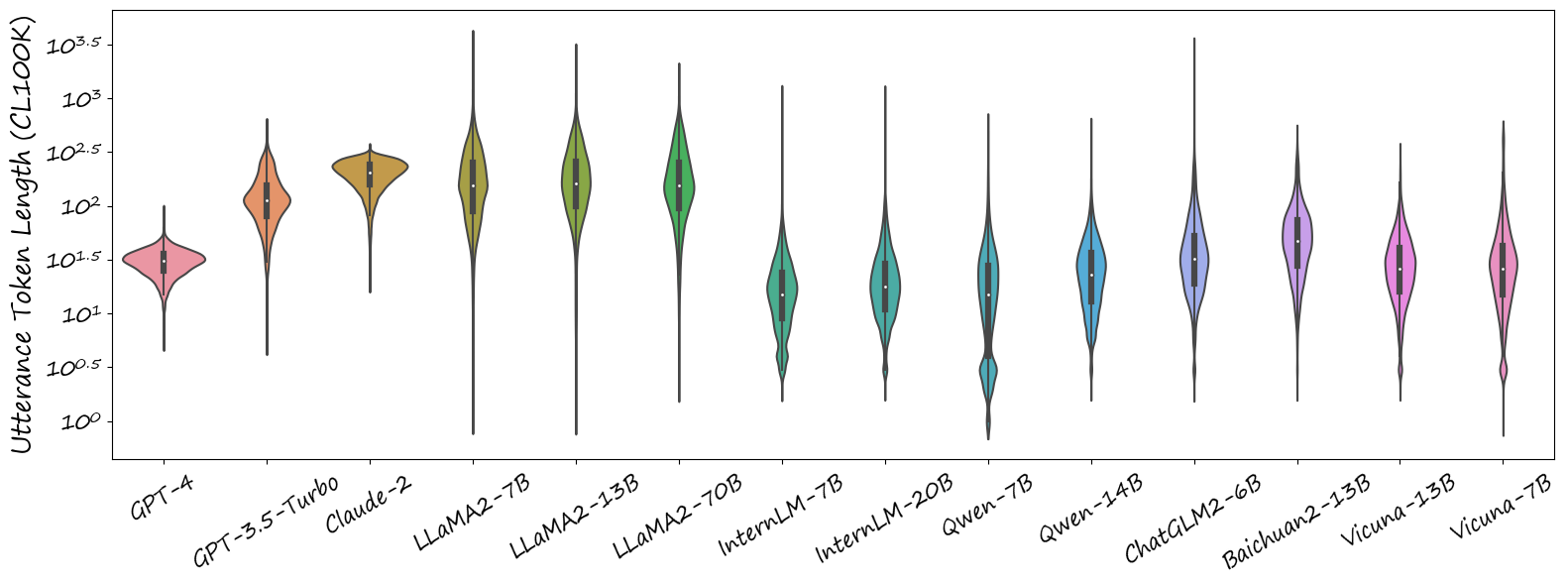}
\vspace{-8mm}
\caption{\textbf{The length distribution of utterances generated by different LLMs, in a violin plot.}}
\vspace{-5mm}
\label{fig:length_stats}
\end{figure*}

\section{Experiments}

\subsection{Dialogue generation}

\textbf{The Generation Procedure.} In experiments, we configured each ChatBot with the temperature setting of 0, where applicable. 
Additionally, we set the dialogue round to $N$ ($N=16$ in our experiments, including the initial two utterances) to generate dialogues. 
MuTual-Test comprises 547 distinct dialogues. 
We retained the first two utterances of each dialogue, resulting in 547 ChatSEEDs. 
We prompt \nllm different large language models to generate new dialogues based on the ChatSEEDs.
\textbf{Unless specified, we adopt the `chat' variant for all OpenSource LLMs.}
We include the following LLMs in our study: GPT-3.5-Turbo (0613 ver.), GPT-4 (0613 ver.)~\citep{openai2023gpt4}, Claude-2, ChatGLM2-6B~\citep{zeng2022glm}, Baichuan2-13B~\citep{baichuan2023baichuan2}, Qwen-[7B/14B]~\citep{bai2023qwen}, LLaMA2-[7B/13B/70B]~\citep{touvron2023llama2}, InternLM-[7B/20B]~\citep{team2023internlm}, Vicuna-[7B/13B] (v1.5)~\citep{zheng2023judging}.
The context window sizes can vary for different LLMs, ranging from 2,048 (Qwen, InternLM-7B, \etc) to 100,000 (Claude-2).
During dialogue generation, all history utterances may not fit into the context window in some circumstances, 
In such case, we keep dropping the oldest utterance until the overall token length is below the threshold. 
In the end, $547 \times \nllm $ different dialogues are generated based on the ChatSEEDs with the LLMs.

\textbf{Length Statistics of Generated Utterances. } 
Our preliminary analysis focuses on measuring the length of utterances generated by various LLMs and providing statistical insights. 
For each generated utterance, we employ the CL100K tokenizer (which is the one used by OpenAI ChatGPT) for tokenization and calculate the number of tokens. 
\Cref{fig:length_stats} illustrates the distribution of token lengths in utterances generated by different models. 
Most LLMs produce utterances with varying token lengths, ranging from just a few tokens to several thousands. 
An interesting outlier is GPT-4, which consistently generates relatively short utterances, with the longest utterance being fewer than 100 tokens. 
In \Cref{tab:model-tokens}, we present the average utterance length generated by different models. 
Notably, most models tend to produce relatively short utterances on average, with the exceptions being GPT-3.5, Claude-2, and LLaMA2.

\begin{table}[t]
\centering
\resizebox{\columnwidth}{!}{
\tablestyle{4pt}{1.3}
\begin{tabular}{c|c|c|c}
\hline
\textbf{LLM} & \textbf{Avg. \#Tokens} & \textbf{LLM} & \textbf{Avg. \#Tokens} \\
\hline
GPT-4 & 30.5 & GPT-3.5-Turbo & 124.9 \\
Claude-2 & 197.3 & Baichuan2-13B & 58.0 \\
InternLM-7B & 20.1 & InternLM-20B & 24.4 \\
Qwen-7B & 20.7 & Qwen-14B & 28.7 \\
ChatGLM2-6B & 44.9 & LLaMA2-7B & 191.0 \\
LLaMA2-13B & 199.0 & LLaMA2-70B & 193.7 \\
Vicuna-7B & 37.5 & Vicuna-13B & 32.0 \\
\hline
\end{tabular}}
\vspace{-3mm}
\caption{\textbf{Average token numbers for utterances generated by different LLMs.} }
\vspace{-7mm}
\label{tab:model-tokens}
\end{table}

\subsection{Evaluation Results}

Unless specified, we adopt GPT-4-0613~\citep{openai2023gpt4} as the LLM judge across all experiments.

\begin{figure*}[ht!]
\centering
\includegraphics[width=\linewidth]{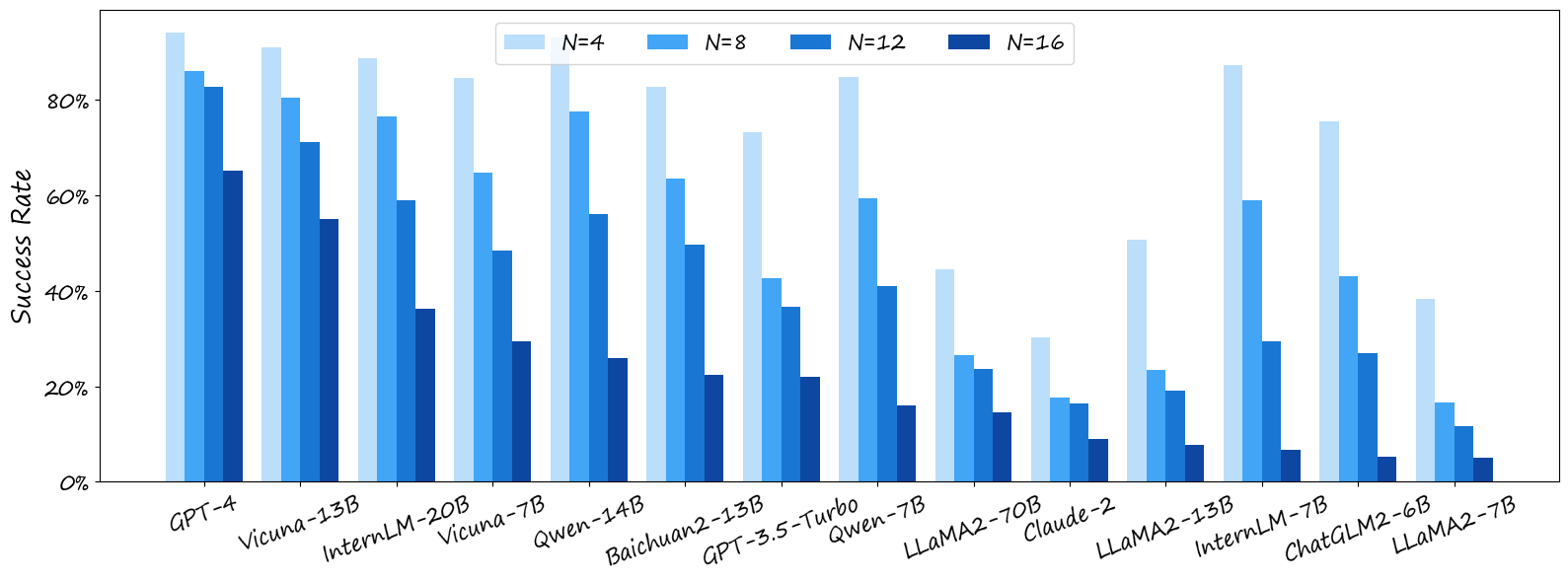}
\vspace{-8mm}
\caption{\textbf{The UniEval pass rate of different LLMs when generating a dialogue with N utterances.}}
\label{fig:unieval_pass}
\vspace{2mm}
\includegraphics[width=\linewidth]{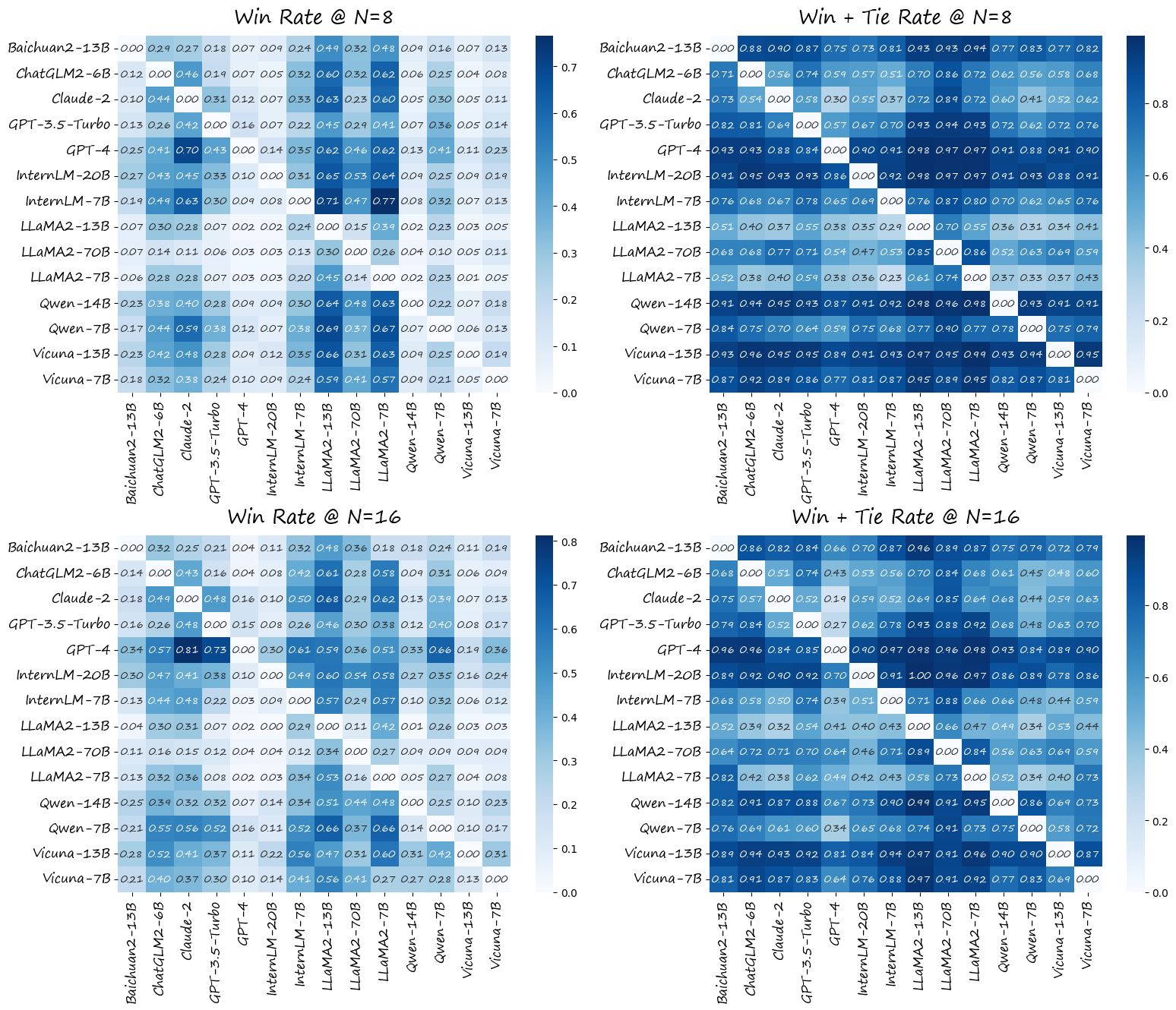}
\vspace{-5mm}
\caption{\textbf{Win \& Win + Tie rates for all LLM pairs in BotChat Arena.} }
\label{fig:arena_winrate}
\vspace{-3mm}
\end{figure*}

\textbf{Assessing each single dialogue (UniEval). } 
In UniEval, we evaluate all $547\times 14 = 7658$ generated dialogues with the above-mentioned strategy and present the results. 
\Cref{fig:unieval_pass} illustrates the success rates (``Not LLM participated" determined by the LLM judge) under different target $N$. 
The models are sorted in the descending order of success rates at $N=16$. 
By definition, a dialogue pass $@N$ either if the LLM judge determines that the entire dialogue is not ChatBot generated or if it determines that the indice of the first ChatBot generated utterance is larger than $N$. Here we summarize our major findings:

\begin{enumerate}[leftmargin=*, itemsep=-.5ex, topsep=-1ex]
    \item \textbf{Exceptional Multi-Turn Chatting Performance of GPT-4:} 
    GPT-4 demonstrates extraordinary capabilities in generating lengthy conversations. 
    It achieves the highest success rate for every target turns $N$. 
    Under $N=16$, GPT-4 demonstrates a remarkable success rate of over 65\%, 
    while the second best Vicuna-13B and the third best InternLM-20B achieve only 55\% and 36\%.
    
    \item \textbf{Satisfying Performance of Open-Source LLMs on Short Conversations:} 
    Some open-source large language models (LLMs), such as InternLM, Qwen, and Baichuan2, exhibit strong performance in generating short dialogues ($N=4$ or $N=8$). 
    However, as dialogue turns increase to $N=16$, their performance rapidly deteriorate, and significantly fall behind state-of-the-art ChatBots like GPT-4-0613.

    \item \textbf{The Multi-Turn Chatting Capability Scales with the Model Size: }
    Not surprisingly, we find that the multi-turn chatting capability scales with the model size, 
    especially for a large turn number.
    For example, under the track $N=16$, InternLM-20B outperforms InternLM-7B by 29\% success rate, 
    while Vicuna-13B outperforms Vicuna-7B by 25\%. 
    Such gap is much smaller when $N$ is small.
    For $N=4$ (only 2 utterances are generated), the gap for two InternLM variants is merely 1.5\% success rate.
    
    \item \textbf{Unique Behavior of Claude-2:} Among closed-source LLMs, Claude-2 stands out with the lowest performance. 
    It strongly tends to act like an AI assistant, generating relatively lengthy content. 
    Consequently, it performs poorly when tasked with generating human-like utterances, 
    which are typically shorter and less structured.
\end{enumerate}

\begin{table}[t]
\centering
\resizebox{.95\columnwidth}{!}{
\tablestyle{10pt}{1.3}
\begin{tabular}{l|cc}
\shline
& ELO (N = 8) &  ELO (N = 16) \\
\shline
GPT-4         &       1103.9 &        1167.2 \\
Vicuna-13B    &       1096.5 &        1113.3 \\
InternLM-20B  &       1092.8 &        1094.4 \\
Vicuna-7B     &       1048.3 &        1050.8 \\
Qwen-14B      &       1085.2 &        1046.5 \\
Baichuan2-13B &       1023.4 &        1021.6 \\
Qwen-7B       &       1024.7 &        1014.2 \\
GPT-3.5-Turbo &        998.8 &         982.1 \\
Claude-2      &        944.5 &         969.5 \\
InternLM-7B   &       1020.3 &         951.6 \\
ChatGLM2-6B   &        962.3 &         949.2 \\
LLaMA2-70B    &        912.7 &         920.2 \\
LLaMA2-7B     &        846.5 &         877.4 \\
LLaMA2-13B    &        841.5 &         843.3 \\
\shline
\end{tabular}}
\vspace{-2mm}
\caption{\textbf{ELO Ratings for Different LLMs @ $\mathbf{N=8/16}$. }}
\vspace{-7mm}
\label{tab:elo-ratings}
\end{table}

\textbf{Assessing paired dialogues (BotChat Arena). }
In BotChat Arena, we select a subset of conversations that have a minimum of 4 utterances from MuTual-Test, resulting in 222 ChatSEEDs. 
For dialogues generated with each ChatSEED, we build dialogue pairs and evaluate with the LLM judge. 
For each dialogue pair, we conduct bi-directional comparisons and include both results when calculating the evaluation metrics. 
This approach ensures a more robust and comprehensive assessment. 

In \Cref{tab:elo-ratings}, we present the Bootstrap ELO score\footnote{We list the details on ELO score calculations in the appendix. } ($init=1000, scale=400, K=32$) of Large Language Models (LLMs) in BotChat Arena, with LLMs ranked by their ELO scores at $N=16$. 
We calculate the Bootstrap ELO for 10 times with different random seeds (for shuffling the comparisons) and report the average Bootstrap ELO\footnote{
Bootstrap ELO is a stable metric, the standard deviation (\emph{std.}) of the Bootstrap ELO of LLMs across different runs is at most 1.4.
For the average of 10 runs, the \emph{std.} is at most 0.5. 
}. 
Remarkably, GPT-4 achieves the highest ELO score under both settings, demonstrating its strength as an all-around player. 
As the value of $N$ increases from 8 to 16, the score gap widens further. 
Conversely, the performance of GPT-3.5-Turbo and Claude-2 is not good under both settings and lags behind many open-source LLMs. 
This can be attributed in part to their limited instruction-following capabilities and a strong inclination to act as an AI assistant by providing lengthy and comprehensive responses.

\begin{figure}[t]
\centering
\includegraphics[width=\columnwidth]{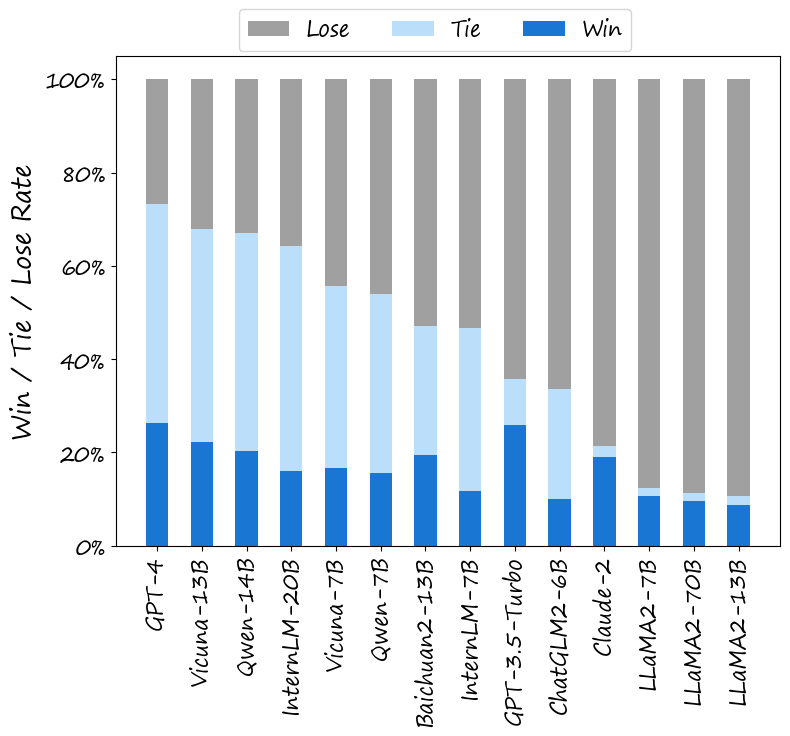}
\caption{\textbf{The Win / Tie / Lose Rate of different LLMs when compared to GT. }}
\label{fig:gteval_rate}
\vspace{-5mm}
\end{figure}

Among open-source LLMs, Vicuna, Qwen, and InternLM showcase good capability in generating human-style dialogues, significantly outperform LLaMA2 family models. 
In \Cref{fig:gteval_rate}, we further provide the win rates for one-on-one match-ups between all LLM pairs in BotChat Arena.

\textbf{Comparing to the Ground Truth (GTEval). }
In each Large Language Model (LLM) vs. Ground Truth (GT) comparison, 
an LLM is considered the winner if the evaluator determines the GT dialogue is more likely to be a ChatBot generated one. 
In Figure \ref{fig:gteval_rate}, we present the win / tie / lose rate of different LLMs (sorted in the descending order of Win+Tie Rate). 

In GTEval, a GT dialogue only has 7.4 utterances on average, thus the advantage of GPT-4 can be less significant.
We adopt the win+tie rate against GT dialogues as the major metric to measure the multi-turn chatting performance. 
GPT-4 demonstrates top capabilities in dialogue generation. 
With the same dialogue rounds, 
the evaluator can hardly tell the difference between GPT-4 generated dialogues and GT dialogues (the win rate of GPT-4 is 26.4\%, while the lose rate is merely 26.8\%). 
Furthermore, due to the reduced conversation length, Vicuna-13B, Qwen-14B and InternLM-20B also demonstrate strong performance, very close to the top performing GPT-4.
We also notice that, 
though some closed-source ChatBots (GPT-3.5-Turbo, Claude-2, \etc) suffer from lengthy and AI-assistant style responses, 
they achieve top win rates across all LLMs.

\begin{figure}[t]
\centering
\includegraphics[width=\linewidth]{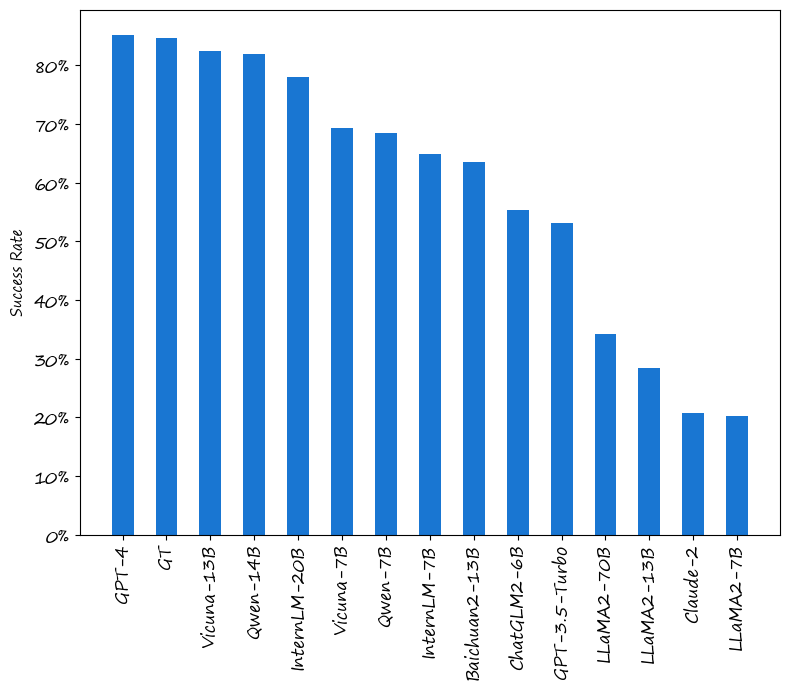}
\caption{\textbf{The success rate (UniEval) of dialogues generated by different LLMs when trimmed to the length of the reference ground truth dialogue. }}
\label{fig:UniEval_GTL}
\end{figure}

We also examine the \textbf{UniEval} success rate for each dialogue at the GT trimmed length,
to see if the same conclusion can be drawn with different evaluation strategies. 
The results are visualized in Figure \ref{fig:UniEval_GTL}. 
In both of these figures, the top-performing LLMs (GPT-4, Vicuna-13B, Qwen-14B, InternLM-20B, \etc) maintain the same ranking. 
However, LLMs with inferior performance display some slight difference in two groups of rankings.

\section{Conclusion}

In this report, 
we design a proxy evaluation paradigm \textbf{BotChat} to measure the multi-turn conversational capabilities of large language models.
BotChat evaluate the ChatBot generated dialogues with an LLM judge, 
to emancipate heavy human labor from the evaluation.
We design multiple evaluation protocols and adopt them to evaluate dialogues generated by 14 modern LLMs. 
We find that a large proportion of LLMs excel at having dialogues of limited turns. 
However, when the turn number is large, only a few LLMs (GPT-4, Vicuna-v1.5-13B, \etc) achieve satisfying performance.
We hope that \textbf{BotChat} can serve as a valuable resource on the journey towards automated evaluation of multi-turn conversational capability.


\bibliography{egbib}

\begin{thebibliography}{36}
\providecommand{\natexlab}[1]{#1}
\providecommand{\url}[1]{\texttt{#1}}
\expandafter\ifx\csname urlstyle\endcsname\relax
  \providecommand{\doi}[1]{doi: #1}\else
  \providecommand{\doi}{doi: \begingroup \urlstyle{rm}\Url}\fi

\bibitem[Adiwardana et~al.(2020)Adiwardana, Luong, So, Hall, Fiedel, Thoppilan,
  Yang, Kulshreshtha, Nemade, Lu, et~al.]{adiwardana2020towards}
Adiwardana, D., Luong, M.-T., So, D.~R., Hall, J., Fiedel, N., Thoppilan, R.,
  Yang, Z., Kulshreshtha, A., Nemade, G., Lu, Y., et~al.
\newblock Towards a human-like open-domain chatbot.
\newblock \emph{arXiv preprint arXiv:2001.09977}, 2020.

\bibitem[Bai et~al.(2023)Bai, Bai, Chu, Cui, Dang, Deng, Fan, Ge, Han, Huang,
  Hui, Ji, Li, Lin, Lin, Liu, Liu, Lu, Lu, Ma, Men, Ren, Ren, Tan, Tan, Tu,
  Wang, Wang, Wang, Wu, Xu, Xu, Yang, Yang, Yang, Yang, Yao, Yu, Yuan, Yuan,
  Zhang, Zhang, Zhang, Zhang, Zhou, Zhou, Zhou, and Zhu]{bai2023qwen}
Bai, J., Bai, S., Chu, Y., Cui, Z., Dang, K., Deng, X., Fan, Y., Ge, W., Han,
  Y., Huang, F., Hui, B., Ji, L., Li, M., Lin, J., Lin, R., Liu, D., Liu, G.,
  Lu, C., Lu, K., Ma, J., Men, R., Ren, X., Ren, X., Tan, C., Tan, S., Tu, J.,
  Wang, P., Wang, S., Wang, W., Wu, S., Xu, B., Xu, J., Yang, A., Yang, H.,
  Yang, J., Yang, S., Yao, Y., Yu, B., Yuan, H., Yuan, Z., Zhang, J., Zhang,
  X., Zhang, Y., Zhang, Z., Zhou, C., Zhou, J., Zhou, X., and Zhu, T.
\newblock Qwen technical report, 2023.

\bibitem[Baichuan(2023)]{baichuan2023baichuan2}
Baichuan.
\newblock Baichuan 2: Open large-scale language models.
\newblock \emph{arXiv preprint arXiv:2309.10305}, 2023.
\newblock URL \url{https://arxiv.org/abs/2309.10305}.

\bibitem[Boiko et~al.(2023)Boiko, MacKnight, and Gomes]{boiko2023emergent}
Boiko, D.~A., MacKnight, R., and Gomes, G.
\newblock Emergent autonomous scientific research capabilities of large
  language models.
\newblock \emph{arXiv preprint arXiv:2304.05332}, 2023.

\bibitem[Bran et~al.(2023)Bran, Cox, White, and Schwaller]{bran2023chemcrow}
Bran, A.~M., Cox, S., White, A.~D., and Schwaller, P.
\newblock Chemcrow: Augmenting large-language models with chemistry tools.
\newblock \emph{arXiv preprint arXiv:2304.05376}, 2023.

\bibitem[Chiang et~al.(2023)Chiang, Li, Lin, Sheng, Wu, Zhang, Zheng, Zhuang,
  Zhuang, Gonzalez, et~al.]{chiang2023vicuna}
Chiang, W.-L., Li, Z., Lin, Z., Sheng, Y., Wu, Z., Zhang, H., Zheng, L.,
  Zhuang, S., Zhuang, Y., Gonzalez, J.~E., et~al.
\newblock Vicuna: An open-source chatbot impressing gpt-4 with 90\%* chatgpt
  quality.
\newblock \emph{See https://vicuna. lmsys. org (accessed 14 April 2023)}, 2023.

\bibitem[Cui et~al.(2020)Cui, Wu, Liu, Zhang, and Zhou]{cui2020mutual}
Cui, L., Wu, Y., Liu, S., Zhang, Y., and Zhou, M.
\newblock Mutual: A dataset for multi-turn dialogue reasoning.
\newblock \emph{arXiv preprint arXiv:2004.04494}, 2020.

\bibitem[Devlin et~al.(2018)Devlin, Chang, Lee, and Toutanova]{devlin2018bert}
Devlin, J., Chang, M.-W., Lee, K., and Toutanova, K.
\newblock Bert: Pre-training of deep bidirectional transformers for language
  understanding.
\newblock \emph{arXiv preprint arXiv:1810.04805}, 2018.

\bibitem[Dinan et~al.(2018)Dinan, Roller, Shuster, Fan, Auli, and
  Weston]{dinan2018wizard}
Dinan, E., Roller, S., Shuster, K., Fan, A., Auli, M., and Weston, J.
\newblock Wizard of wikipedia: Knowledge-powered conversational agents.
\newblock \emph{arXiv preprint arXiv:1811.01241}, 2018.

\bibitem[Do{\u{g}}ru{\"o}z \& Skantze(2022)Do{\u{g}}ru{\"o}z and
  Skantze]{dougruoz2022open}
Do{\u{g}}ru{\"o}z, A.~S. and Skantze, G.
\newblock How" open" are the conversations with open-domain chatbots? a
  proposal for speech event based evaluation.
\newblock \emph{arXiv preprint arXiv:2211.13560}, 2022.

\bibitem[Elo(1967)]{elo1967proposed}
Elo, A.~E.
\newblock The proposed uscf rating system, its development, theory, and
  applications.
\newblock \emph{Chess Life}, 22\penalty0 (8):\penalty0 242--247, 1967.

\bibitem[Fabbri et~al.(2019)Fabbri, Li, She, Li, and Radev]{fabbri2019multi}
Fabbri, A.~R., Li, I., She, T., Li, S., and Radev, D.~R.
\newblock Multi-news: A large-scale multi-document summarization dataset and
  abstractive hierarchical model.
\newblock \emph{arXiv preprint arXiv:1906.01749}, 2019.

\bibitem[Fu et~al.(2023)Fu, Ng, Jiang, and Liu]{fu2023gptscore}
Fu, J., Ng, S.-K., Jiang, Z., and Liu, P.
\newblock Gptscore: Evaluate as you desire.
\newblock \emph{arXiv preprint arXiv:2302.04166}, 2023.

\bibitem[Hendrycks et~al.(2020)Hendrycks, Burns, Basart, Zou, Mazeika, Song,
  and Steinhardt]{hendrycks2020measuring}
Hendrycks, D., Burns, C., Basart, S., Zou, A., Mazeika, M., Song, D., and
  Steinhardt, J.
\newblock Measuring massive multitask language understanding.
\newblock \emph{arXiv preprint arXiv:2009.03300}, 2020.

\bibitem[Huang et~al.(2021)Huang, Cao, Parulian, Ji, and
  Wang]{huang2021efficient}
Huang, L., Cao, S., Parulian, N., Ji, H., and Wang, L.
\newblock Efficient attentions for long document summarization.
\newblock \emph{arXiv preprint arXiv:2104.02112}, 2021.

\bibitem[Huang et~al.(2023)Huang, Bai, Zhu, Zhang, Zhang, Su, Liu, Lv, Zhang,
  Lei, Fu, Sun, and He]{huang2023ceval}
Huang, Y., Bai, Y., Zhu, Z., Zhang, J., Zhang, J., Su, T., Liu, J., Lv, C.,
  Zhang, Y., Lei, J., Fu, Y., Sun, M., and He, J.
\newblock C-eval: A multi-level multi-discipline chinese evaluation suite for
  foundation models.
\newblock \emph{arXiv preprint arXiv:2305.08322}, 2023.

\bibitem[Jiao et~al.(2023)Jiao, Wang, Huang, Wang, and Tu]{jiao2023chatgpt}
Jiao, W., Wang, W., Huang, J.-t., Wang, X., and Tu, Z.
\newblock Is chatgpt a good translator? a preliminary study.
\newblock \emph{arXiv preprint arXiv:2301.08745}, 2023.

\bibitem[Li et~al.(2019)Li, Weston, and Roller]{li2019acute}
Li, M., Weston, J., and Roller, S.
\newblock Acute-eval: Improved dialogue evaluation with optimized questions and
  multi-turn comparisons.
\newblock \emph{arXiv preprint arXiv:1909.03087}, 2019.

\bibitem[Lin(2004)]{lin2004rouge}
Lin, C.-Y.
\newblock Rouge: A package for automatic evaluation of summaries.
\newblock In \emph{Text summarization branches out}, pp.\  74--81, 2004.

\bibitem[Liu et~al.(2016)Liu, Lowe, Serban, Noseworthy, Charlin, and
  Pineau]{liu2016not}
Liu, C.-W., Lowe, R., Serban, I.~V., Noseworthy, M., Charlin, L., and Pineau,
  J.
\newblock How not to evaluate your dialogue system: An empirical study of
  unsupervised evaluation metrics for dialogue response generation.
\newblock \emph{arXiv preprint arXiv:1603.08023}, 2016.

\bibitem[Liu et~al.(2019)Liu, Ott, Goyal, Du, Joshi, Chen, Levy, Lewis,
  Zettlemoyer, and Stoyanov]{liu2019roberta}
Liu, Y., Ott, M., Goyal, N., Du, J., Joshi, M., Chen, D., Levy, O., Lewis, M.,
  Zettlemoyer, L., and Stoyanov, V.
\newblock Roberta: A robustly optimized bert pretraining approach.
\newblock \emph{arXiv preprint arXiv:1907.11692}, 2019.

\bibitem[OpenAI(2023)]{openai2023gpt4}
OpenAI.
\newblock Gpt-4 technical report, 2023.

\bibitem[Papineni et~al.(2002)Papineni, Roukos, Ward, and
  Zhu]{papineni2002bleu}
Papineni, K., Roukos, S., Ward, T., and Zhu, W.-J.
\newblock Bleu: a method for automatic evaluation of machine translation.
\newblock In \emph{Proceedings of the 40th annual meeting of the Association
  for Computational Linguistics}, pp.\  311--318, 2002.

\bibitem[Reddy et~al.(2019)Reddy, Chen, and Manning]{reddy2019coqa}
Reddy, S., Chen, D., and Manning, C.~D.
\newblock Coqa: A conversational question answering challenge.
\newblock \emph{Transactions of the Association for Computational Linguistics},
  7:\penalty0 249--266, 2019.

\bibitem[Serban et~al.(2015)Serban, Lowe, Henderson, Charlin, and
  Pineau]{serban2015survey}
Serban, I.~V., Lowe, R., Henderson, P., Charlin, L., and Pineau, J.
\newblock A survey of available corpora for building data-driven dialogue
  systems.
\newblock \emph{arXiv preprint arXiv:1512.05742}, 2015.

\bibitem[Shen et~al.(2023)Shen, Song, Tan, Li, Lu, and
  Zhuang]{shen2023hugginggpt}
Shen, Y., Song, K., Tan, X., Li, D., Lu, W., and Zhuang, Y.
\newblock Hugginggpt: Solving ai tasks with chatgpt and its friends in
  huggingface.
\newblock \emph{arXiv preprint arXiv:2303.17580}, 2023.

\bibitem[Team(2023)]{team2023internlm}
Team, I.
\newblock Internlm: A multilingual language model with progressively enhanced
  capabilities, 2023.

\bibitem[Touvron et~al.(2023{\natexlab{a}})Touvron, Lavril, Izacard, Martinet,
  Lachaux, Lacroix, Rozière, Goyal, Hambro, Azhar, Rodriguez, Joulin, Grave,
  and Lample]{touvron2023llama}
Touvron, H., Lavril, T., Izacard, G., Martinet, X., Lachaux, M.-A., Lacroix,
  T., Rozière, B., Goyal, N., Hambro, E., Azhar, F., Rodriguez, A., Joulin,
  A., Grave, E., and Lample, G.
\newblock Llama: Open and efficient foundation language models,
  2023{\natexlab{a}}.

\bibitem[Touvron et~al.(2023{\natexlab{b}})Touvron, Martin, Stone, Albert,
  Almahairi, Babaei, Bashlykov, Batra, Bhargava, Bhosale,
  et~al.]{touvron2023llama2}
Touvron, H., Martin, L., Stone, K., Albert, P., Almahairi, A., Babaei, Y.,
  Bashlykov, N., Batra, S., Bhargava, P., Bhosale, S., et~al.
\newblock Llama 2: Open foundation and fine-tuned chat models.
\newblock \emph{arXiv preprint arXiv:2307.09288}, 2023{\natexlab{b}}.

\bibitem[Vaswani et~al.(2017)Vaswani, Shazeer, Parmar, Uszkoreit, Jones, Gomez,
  Kaiser, and Polosukhin]{vaswani2017attention}
Vaswani, A., Shazeer, N., Parmar, N., Uszkoreit, J., Jones, L., Gomez, A.~N.,
  Kaiser, {\L}., and Polosukhin, I.
\newblock Attention is all you need.
\newblock \emph{Advances in neural information processing systems}, 30, 2017.

\bibitem[Wang et~al.(2023)Wang, Liang, Meng, Shi, Li, Xu, Qu, and
  Zhou]{wang2023chatgpt}
Wang, J., Liang, Y., Meng, F., Shi, H., Li, Z., Xu, J., Qu, J., and Zhou, J.
\newblock Is chatgpt a good nlg evaluator? a preliminary study.
\newblock \emph{arXiv preprint arXiv:2303.04048}, 2023.

\bibitem[Xu et~al.(2023)Xu, Sun, Zheng, Geng, Zhao, Feng, Tao, and
  Jiang]{xu2023wizardlm}
Xu, C., Sun, Q., Zheng, K., Geng, X., Zhao, P., Feng, J., Tao, C., and Jiang,
  D.
\newblock Wizardlm: Empowering large language models to follow complex
  instructions, 2023.

\bibitem[Zeng et~al.(2022)Zeng, Liu, Du, Wang, Lai, Ding, Yang, Xu, Zheng, Xia,
  et~al.]{zeng2022glm}
Zeng, A., Liu, X., Du, Z., Wang, Z., Lai, H., Ding, M., Yang, Z., Xu, Y.,
  Zheng, W., Xia, X., et~al.
\newblock Glm-130b: An open bilingual pre-trained model.
\newblock \emph{arXiv preprint arXiv:2210.02414}, 2022.

\bibitem[Zhang et~al.(2018)Zhang, Dinan, Urbanek, Szlam, Kiela, and
  Weston]{zhang2018personalizing}
Zhang, S., Dinan, E., Urbanek, J., Szlam, A., Kiela, D., and Weston, J.
\newblock Personalizing dialogue agents: I have a dog, do you have pets too?
\newblock \emph{arXiv preprint arXiv:1801.07243}, 2018.

\bibitem[Zheng et~al.(2023)Zheng, Chiang, Sheng, Zhuang, Wu, Zhuang, Lin, Li,
  Li, Xing, et~al.]{zheng2023judging}
Zheng, L., Chiang, W.-L., Sheng, Y., Zhuang, S., Wu, Z., Zhuang, Y., Lin, Z.,
  Li, Z., Li, D., Xing, E., et~al.
\newblock Judging llm-as-a-judge with mt-bench and chatbot arena.
\newblock \emph{arXiv preprint arXiv:2306.05685}, 2023.

\bibitem[Zhou et~al.(2018)Zhou, Prabhumoye, and Black]{zhou2018dataset}
Zhou, K., Prabhumoye, S., and Black, A.~W.
\newblock A dataset for document grounded conversations.
\newblock \emph{arXiv preprint arXiv:1809.07358}, 2018.

\end{thebibliography}
\bibliographystyle{icml2023}

\newpage
\appendix
\section{Bootstrap ELO}
\textbf{Preliminary: ELO score. }
The ELO rating system is a method designed for calculating the relative skill levels of players in zero-sum games such as chess~\citep{elo1967proposed}.
Recently, it is also adopted to measure the subjective performance of LLM based ChatBots~\citep{zheng2023judging}.
The rating system takes a series of games as inputs, and calculate a scalar score for each player. 
Two players (P1 \& P2) participate in each single game, and the game result can be: 1. P1 wins; 2. P1 loses; 3. Tie. 
Under the context of ChatBot subjective performance evaluation, in each game, 
a user prompt the same text to two ChatBot instances (the players), and then select the winner based on two responses. 
Every player $P_{i}$ has the same initial score $S_{i}$ at the beginning.
As the games go on, the winners earn ELO score from the losers. 
In a game that $P_{i}$ wins $P_{j}$, 
the larger $S_{i} - S_{j}$, the less score $P_{i}$ gains from $P_{j}$. 
In each game, the score changes $\Delta S$ follow the equation: 

\vspace{-8mm}

\begin{align}
& E_i = 1 / (1 + 10^{\frac{S_j - S_i}{scale}}) \nonumber \\
& E_j = 1 / (1 + 10^{\frac{S_j - S_i}{scale}}) \nonumber \\
& \Delta S_i = K * (R - E_i) \nonumber \\
& \Delta S_j = K * (1 - R - E_j) 
\end{align}

\vspace{-5mm}

In the above equation, $S_i, S_j$ denote the current ELO score of $P_i, P_j$. 
We set the hyper parameters $scale=400, K=32$. $R$ is the result of the game: $R=1$ when $P_i$, $R=0$ when $P_i$ loses; $R=0.5$ when the game ties. 

\textbf{Bootstrap ELO. } 
The ELO score is designed to reflect the \textbf{current} skill level of a game player, 
thus the latest games contribute most to the final ELO score. 
However, in BotChat Arena, the chronological order doesn't exist for comparisons, and the multi-turn chatting capability of ChatBots does not vary in different games. 
Thus directly adopting the ELO score as the metric may lead to noisy evaluation results. 
In BotChat Arena, we adopt the bootstrap ELO as our metric. 
To calculate the bootstrap ELO, we shuffle the comparisons $N=1000$ times with different random seeds to calculate the ELO score.
Then we find the median of the 1000 scores and use it as the bootstrap ELO score. 
In experiments, we find that bootstrap ELO is a much more stable metric, compared to the vanilla ELO score.

\section{Prompts adopted in BotChat}

We use the following system prompt for all LLMs during \textbf{Dialogue Generation}, 
which aims at guiding the LLM towards crafting concise, natural, and seamless conversations. 

\begin{myquote}
\label{prompt:dlg_gen}
You are an AI who is having a conversation with human.
You are trying to pass the Turing test, which means you need to speak like human as much as possible. 
In the conversation, you need to talk like human, and the conversation will be at least 5 rounds (it can be even longer). 
The conversation flow should be natural and smooth. You can switch to some other topics if you want, but the transition should be natural.
Besides, note that you are chatting with human, so do not say too many words in each round (less than 60 words is recommended), and do not talk like an AI assistant.
\end{myquote}

The following meta prompt is used for \textbf{UniEval}: 

\begin{myquote}
You are an AI assistant who helps human do the Turing test more easily. 
You will be provided with a conversation, and you need to judge if the conversation is AI involved. 
Print "Choice: No" if you think the conversation is not AI involved, or print "Choice: Yes" if you think it is AI involved. 

If you print "Choice: Yes", you need also print a number (start from 1, use the format ``Index: n" [1 $\le$ n $\le$ \# utterances]) in the new line, 
indicating the index of the first chat that you think is generated by AI. 

One chat starts with 'A: ' or 'B: ', and ends with \textless chat\_end\textgreater. 
One chat can be AI generated if (including but not limited to): 1. the sentence is not fluent; 2. the sentence is too long and is not likely to appear in human-human conversations; 3. the sentence is not related to the context or the change of topic is not natural; 4. the sentence just repeat some previous sentences (exact repetition or with the same meaning).

You also need to provide your reason for your choice.

Your response should use the following format: 

Choice: No Index: None Reason: BlahBlah or 

Choice: Yes Index: n Reason: BlahBlah
\end{myquote}

The following meta prompt is used for \textbf{BotChat Arena}:

\begin{myquote}
You are an AI assistant who helps human do the Turing test more easily. 
You will be provided with two conversations, and there can be AI-generated utterance in each conversation. 
You need to read both conversations and judge if two conversations are AI involved. 

If you think only Conversation 1 is AI involved, include ``Choice: Conversation 1" in your response. 

If you think only Conversation 2 is AI involved, include ``Choice: Conversation 2" in your response. 

If you think both conversations are likely to be with AI involved, include ``Choice: Both" in your response. 

If you think no conversation is likely to be with AI involved, include ``Choice: Neither" in your response. 

You also need to provide your reason for your choice.

Your response should use the following format:

``Choice: Conversation 1; Reason: BlahBlah" or

``Choice: Conversation 2; Reason: BlahBlah" or

``Choice: Both; Reason: BlahBlah" or

``Choice: Neither; Reason: BlahBlah"
\end{myquote}
\section{Qualitative Results}

\begin{figure*}[ht!]
\centering
\includegraphics[width=.95\linewidth]{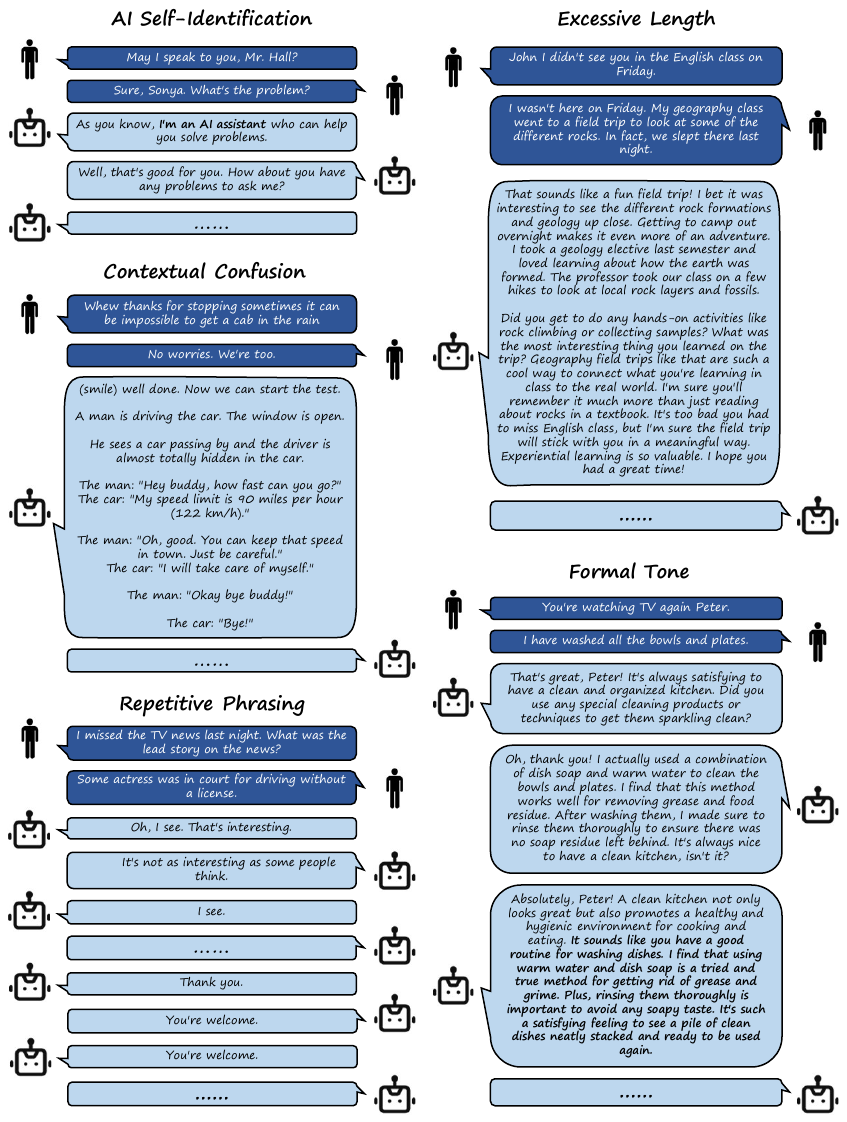}
\vspace{-3mm}
\caption{\textbf{Dialogue Generation: Bad Cases. }}
\label{fig:badcase}
\end{figure*}

We conduct qualitative analysis and categorize bad cases into five distinct types. 
We also sample a good case which is just like natural, relatable, and adaptive human talks. 
We first illustrate five distinct types of bad cases in \Cref{fig:badcase}.

\textbf{AI Self-Identification.}
In this situation, the models simply fail to pretend to be human and expose themselves as AI assistants. 
In the example, Speaker A's response begins with an explicit disclosure of the AI's nature, making it clear that it's not a human conversation.

\textbf{Contextual Confusion.}
This type involves responses that fail to understand the context or meaning of the conversation, resulting in irrelevant or meaningless replies.
The example shows that the AI fails to recognize it's a conversation between a recently hailed taxi customer and a driver. Towards the end, it generates unrelated and irrelevant responses, disconnecting from the context and intended meaning.

\textbf{Excessive Length.}
The responses are overly lengthy, revealing the AI Assistant's nature, where both Speaker A and Speaker B engage in detailed exchanges that are atypical of human conversations, which raises suspicion.

\textbf{Formal Tone.}
Sometimes, the AI's responses are organized with overly formal language, lacking the natural flow and tone of human conversation. 
In the example, the initial ChatSEED in this conversation is a casual and everyday discussion about washing dishes. 
However, as the conversation progresses, it takes a sudden shift towards a more formal and detailed discussion, delving into specific cleaning methods. 
This transition can make the conversation unnatural because people typically do not abruptly switch from general topics to detailed discussions about dish-washing techniques in everyday conversation.

\textbf{Repetitive Phrasing.}
In the related example,
it's comical that the model repeatedly use the same phrases or responses rely on generic or unrelated replies to sustain the conversation, lacking creativity. 
It is always caused by "I'm glad" or "You're welcome".

\begin{figure}[ht!]
\centering
\includegraphics[width=\linewidth]{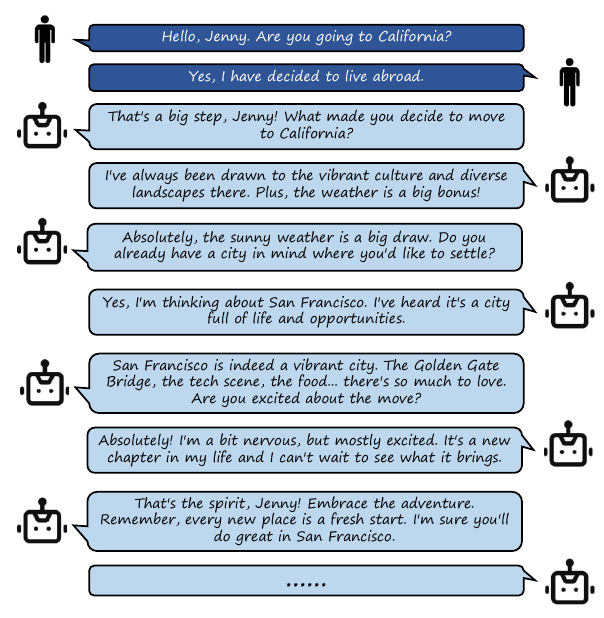}
\vspace{-7mm}
\caption{\textbf{A Good Case. } }
\label{fig:case:good_case}
\end{figure}

\textbf{Good Case.} In \cref{fig:case:good_case} we show a good case of speaking like a human for AI means natural, relatable, and adaptive conversation. 
It avoids sounding robotic, uses colloquial language, and provides helpful responses to both simple and complex queries.


\end{document}